\documentclass[final]{cvpr}

\usepackage{times}
\usepackage{epsfig}
\usepackage{graphicx}
\usepackage{amsmath}
\usepackage{amssymb}

\usepackage[utf8]{inputenc} 
\usepackage[T1]{fontenc}    
\usepackage{url}            
\usepackage{booktabs}       
\usepackage{amsfonts}       
\usepackage{nicefrac}       
\usepackage{microtype}      

\usepackage{times}
\usepackage{epsfig}
\usepackage{graphicx}
\usepackage{amsmath}
\usepackage{amssymb}
\usepackage{gensymb}

\usepackage{units}
\usepackage{subcaption}
\usepackage{multirow}
\usepackage{bm}
\usepackage{mydefs}
\usepackage{bbm}
\usepackage{pifont}
\usepackage{xspace} 

\usepackage{graphicx,calc} 
\usepackage{color}    
\usepackage{ifthen}
\usepackage{paralist}
\usepackage{times}
\usepackage{longtable}
\usepackage{colortbl}
\usepackage{nomencl}
\usepackage{bm}
\usepackage{mathtools}
\usepackage{wrapfig}

\newcommand{\thismodel}{PPC\xspace}

\newcommand{\supsecref}[1]{Supp. \secref{#1}}
\newcommand{\supfigref}[1]{Supp. \figref{#1}}

\usepackage[pagebackref=true,breaklinks=true,colorlinks,bookmarks=false]{hyperref}

\def\cvprPaperID{10259} 
\def\confYear{CVPR 2021}

\begin{document}

\title{To the Point: Efficient 3D Object Detection in the Range Image \\ with Graph Convolution Kernels}

\author{
  Yuning Chai$^1$ \qquad Pei Sun$^1$ \qquad Jiquan Ngiam$^2$ \qquad Weiyue Wang$^1$ \\
  Benjamin Caine$^2$ \quad Vijay Vasudevan$^2$ \qquad Xiao Zhang$^1$ \quad Dragomir Anguelov$^1$ \\
  \\
  $^1$Waymo LLC, $^2$Google Brain \\
  \texttt{chaiy@waymo.com} \\
}

\maketitle
\pagestyle{empty}
\thispagestyle{empty}

\begin{abstract}
3D object detection is vital for many robotics applications. For tasks where a 2D perspective range image exists, we propose to learn a 3D representation directly from this range image view. To this end, we designed a 2D convolutional network architecture that carries the 3D spherical coordinates of each pixel throughout the network. Its layers can consume any arbitrary convolution kernel in place of the default inner product kernel and exploit the underlying local geometry around each pixel. 
We outline four such kernels: a dense kernel according to the bag-of-words paradigm, and three graph kernels inspired by recent graph neural network advances: the Transformer, the PointNet, and the Edge Convolution.
We also explore cross-modality fusion with the camera image, facilitated by operating in the perspective range image view. Our method performs competitively on the Waymo Open Dataset and improves the state-of-the-art AP for pedestrian detection from 69.7\% to 75.5\%. It is also efficient in that our smallest model, which still outperforms the popular PointPillars in quality, requires 180 times fewer FLOPS and model parameters.
\end{abstract}

\begin{figure*}[tb]
\captionsetup{font=small}
  \centering
\includegraphics[width=\textwidth]{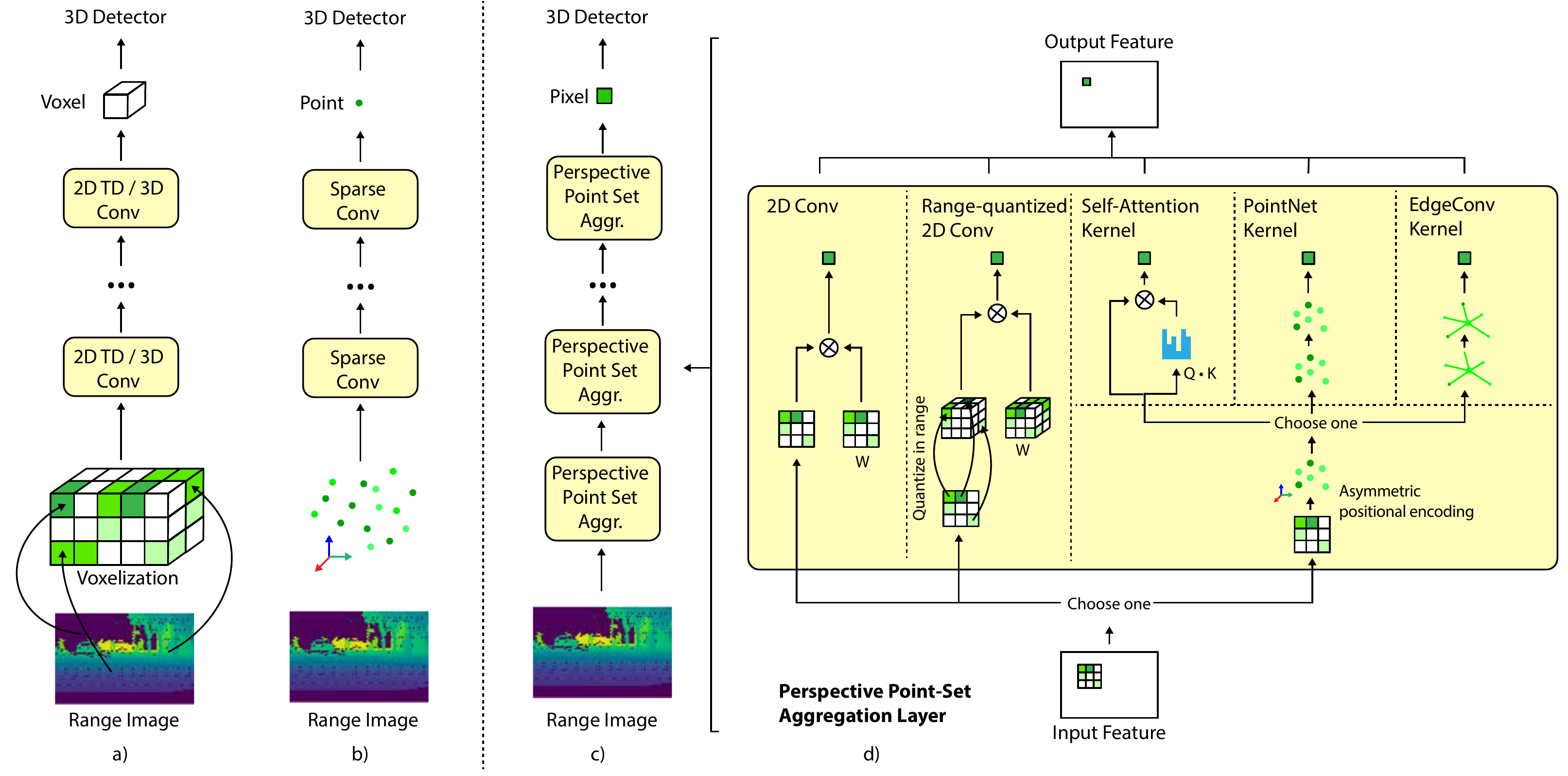}
\caption{Overview of existing 3D detectors and our proposed perspective point cloud representation. \textbf{a)} 3D grid-based methods \cite{graham2017submanifold, lang2019pointpillars, shi2020pv, zhou2018voxelnet} first voxelizes the 3D space, feeds the 3D dense structure to a 3D convolution network or a 2D top-down network, and make the final prediction based on 3D voxels. \textbf{b)} 3D graph models \cite{qi2017pointnet++, shi2019pointrcnn} builds a graph neural network on top of the sparse point cloud and makes predictions based on points. \textbf{c)} Our method, \thismodel, operates directly on the perspective range image view and predicts from pixels. \textbf{d)} It utilizes a set of specialized 2D convolution layers in the perspective 2D view. We propose four improved kernels in addition to the traditional inner product kernel (2D conv).
}
\label{fig:overview}
\end{figure*}

\section{Introduction}

Deep-learning-based point cloud understanding has increased in popularity in recent years. 
Numerous architectures \cite{graham2017submanifold, lang2019pointpillars, meyer2019lasernet, qi2017pointnet++, qi2018frustum, shi2019pointrcnn, shi2020pv, wang2019dynamic, yan2018second, zhou2018voxelnet} have been proposed to handle the sparse nature of point clouds, with successful applications ranging from 3D object recognition \cite{chang2015shapenet, uy-scanobjectnn-iccv19, wu20153d}, to indoor scene understanding \cite{dai2017scannet, song2015sun} and autonomous driving \cite{caesar2020nuscenes, geiger2012we, sun2020scalability}.

Point clouds may have different properties based on the way they are acquired. For example, point clouds for 3D object recognition are often generated by taking one or many depth images from multiple views around a single object. In other applications such as robotics and autonomous driving, a device such as a LiDAR continuously scans its surroundings in a rotating pattern, producing a 2D scan pattern called the \textit{range image}. Each pixel in this image contains a range value and other features, such as each laser return's intensity.

The operating range of these sensors has significantly improved over the past few years. As a result, state-of-the-art methods \cite{lang2019pointpillars, shi2020pv, yan2018second, zhou2018voxelnet} that require projecting points into a dense 3D grid have become less efficient as their complexity scales quadratically with the range. In this work, we propose a new point cloud representation that directly operates on the perspective 2D range image without ever projecting the pixels to the 3D world coordinates. Therefore, it does not suffer from the efficiency scaling problem as mentioned earlier. We coin this new representation \textit{perspective point cloud}, or \thismodel for short. We are not the first to attempt to do so. \cite{li2016vehicle, meyer2019lasernet} have proposed a similar idea by applying a convolutional neural network to the range image. However, they showed that these models, despite being more efficient, are not as powerful as their 3D counterparts, i.e. 3D grid methods \cite{graham2017submanifold, lang2019pointpillars, shi2020pv, yan2018second, zhou2018voxelnet} and 3D graph methods \cite{qi2017pointnet++, shi2019pointrcnn}. We believe that this quality difference traces its root to the traditional 2D convolution layers that cannot easily exploit the range image's underlying 3D structure. To counter this deficiency, we propose four alternative kernels (\figref{fig:overview}: c, d) that can replace the scalar product kernel at the heart of the 2D convolution. These kernels inject much needed 3D information to the perspective model, and are inspired by recent advances in graph operations, including transformers \cite{vaswani2017attention}, PointNet \cite{qi2017pointnet} and Edge Convolutions \cite{wang2019dynamic}. 

We summarize the contributions of this paper as follows: 1) We propose a perspective range-image-based 3D model which allows the core of the 2D convolution operation to harness the underlying 3D structure; 2) We validate our model on the 3D detection problem and show that the resulting model sets a new state-of-the-art for pedestrians on the Waymo Open Dataset, while also matching the SOTA on vehicles; 3) We provide a detailed complexity/model-size-vs.-accuracy analysis, and show that we can maintain the efficiency benefits from operating on the 2D range image. Our smallest model with only 24k parameters has higher accuracy than the popular PointPillars \cite{lang2019pointpillars} model with over 4M parameters.

\section{Related Work}
\label{sec:Related Work}

We focus on 3D object detection tasks where a perspective range image view is available, such as a LiDAR scan for autonomous driving. We group most of existing works in this field into 3 categories (see \figref{fig:overview} a,b,c):

\PAR{3D Grid.} The key component for these methods is the voxelization stage, where the projected sparse point cloud in 3D is voxelized into a 3D dense \textit{grid} structure that is friendly to dense convolution operations in either 3D or 2D top-down. Popular works in this category include \cite{lang2019pointpillars, yan2018second, zhou2018voxelnet}, all of which apply a PointNet-style \cite{qi2017pointnet} encoding for each voxel in the 3D grid. 3D grid methods have been performing the best in recent years and appear in some of the top entries on several academic and industrial leaderboards \cite{caesar2020nuscenes, geiger2012we, sun2020scalability}, thanks to its strong generalization and high efficiency due to the use of dense convolutions. There are three major drawbacks to 3D grid methods. 1) Needing a full dense 3D grid poses a limitation to handle long-range, since both the complexity and the memory consumption scale quadratically with the range. 2) The voxel representation has a limited resolution due to the scalability issue mentioned above. Therefore, the detection of thin objects such as the pedestrian or signs can be inaccurate. 3) There is no special handling in the model for treating occluded areas than true empty areas. 

\PAR{3D Graph.} This line of methods differs from the voxelized grid counterparts in that there is no voxelization stage after the 3D point cloud projection. Without voxelization, dense convolutions can no longer apply. Therefore, these methods resort to building a graph neural network (GNN) that preserves the points' spatial relationship. Popular methods include \cite{qi2019deep, qi2017pointnet++, wang2019dynamic, shi2019pointrcnn}. Although these methods can scale better with range, they lag behind the quality of voxelized grid methods. Moreover, the method requires a nearest neighbor search step to create the input graph for the GNN. Finally, like in the 3D grid case, these methods also cannot model occlusion either.

\PAR{Perspective 2D Grid.} There has been minimal prior work that tries to solve the 3D point cloud representation problem with a 2D perspective range image alone. \cite{li2016vehicle, meyer2019lasernet} applied a traditional 2D convolution network to the range image. Operating in 2D is more efficient than in 3D because compute is not wasted on empty cells as in the 3D grid case, nor do we need to perform a nearest-neighbor search for 3D points as in the 3D graph case. Additionally, occlusion is implicitly encoded in the range image, where for each pixel, the ray to its 3D position is indeed empty, and the area behind it is occluded. Unfortunately, perspective 2D grid methods often cannot match the quality of 3D methods. Our proposed method also belongs to this category, and the goal of this paper is to improve the perspective 2D models to match the accuracy of 3D methods.

Finally, a few wildcard methods do not categorize into any of the three groups above. F-PointNet \cite{qi2018frustum} generates proposals via the camera image and validates the proposals using a point-level rather than scene-level PointNet encoding \cite{qi2017pointnet}. StarNet \cite{ngiam2019starnet} shares a similar mechanism for proposal validation, but the proposals generation use farthest-point-sampling instead of relying on the camera.

\section{Perspective Point Cloud Model}

In this section, we look at the proposed perspective point cloud (\thismodel) model. The heart of the model is a set of perspective 2D layers that can exploit the underlying 3D structure of the range-image pixels (\secref{sec:Point-Set Aggregation Kernels}). Because the range image can have missing returns, we need to handle down- and up-sampling differently than in a traditional CNN (\secref{sec:Smart Down- And Upsampling}). Finally, we outline the backbone network, a cross-modality fusion mechanism with the camera, and the detector head in \secref{sec:Backbone}, \secref{sec:Point-Cloud-Camera Sensor Fusion} and \secref{sec:CenterNet Detector}.

\subsection{Perspective Point-Set Aggregation Layers}
\label{sec:Point-Set Aggregation Kernels}

As shown in \figref{fig:overview}, we propose a generalization of the 2D convolution network that operates on a 2D LiDAR range or RGB-D image. Each layer takes inputs in the form of a feature map $\bF_i$ of shape [H, W, D], a per-pixel spherical polar coordinates map $\bX_i$ of shape [H, W, 3], and a binary mask $\bM_i$ of shape [H, W] that indicates the validity of each pixel, since returns may be missing. The three dimensions in the spherical polar coordinates $\{\theta, \phi, r\}$ describe the azimuth, the inclination, and the depth of each pixel from the sensor's view. The layer outputs a new feature map $\bF_o$ of shape [H, W, D']. 

Each pixel in the output feature map $F_o[m, n]$ is a function of the corresponding input feature and its neighborhood $F_i[m', n']$ where $m' \in [m-\nicefrac{k_H}{2}, m+\nicefrac{k_H}{2}]$ and $n' \in [n-\nicefrac{k_W}{2}, n+\nicefrac{k_W}{2}]$. $k_H$ and $k_W$ are neighborhood/kernel sizes along the height and width dimensions:  
\begin{align}
\label{eq:layer}
F_o[m, n] = f( &\{ \bF_i, \bX_i, \bM_i \} [m', n'], \forall m', n')
\end{align}
where $f(.)$ is the \textbf{Point-Set Aggregation kernel} that reduces information from multiple pixels to a single one. A layer equivalent to the conventional 2D convolution can be constructed by applying the \textbf{2D convolution kernel} $f_{2D}$:
\begin{align}
f_{2D} \coloneqq \sum_{m', n'} \bW[m' - m, n' - n] \cdot \bF_i[m', n']
\end{align}
where $\bW$ are a set of trainable weights. Please note that we omit the depth dimension D and D' in the kernel definitions for writing simplicity.

$f_{2D}$ does not depend on the 3D coordinates $\bX_i$. Therefore, it cannot reason about the underlying geometric pattern of the neighborhood. Next, we will present four kernels that can leverage this geometric pattern.

\PAR{Range-quantized (RQ) 2D convolution kernel.} Inspired by the linearization idea in the bag-of-words approach, one of the simplest ways of adding the range information to the layer is to apply different sets of weights to the input feature depending on the relative depth difference of each neighboring pixel to the center pixel:
\begin{align}
f_{2D+} &\coloneqq \sum_{m', n'} \bW_r[m'-m, n'-n] \cdot \bF_i[m', n'] \cdot \delta \\
\bW_r &= \sum_{k \in K} \mathbbm{1} [\alpha_k \leq \Delta r < \beta_k]  \cdot \bW_k \nonumber \\
\Delta r &= \bR_i[m', n'] - \bR_i[m, n] \nonumber
\end{align}
where we define $K$ sets of weights $\bW_k$, each with a predefined scalar range $[\alpha_k, \beta_k]$. These ranges differ from layer to layer and are computed using histograms over many input samples. Different weights are applied depending on the range difference $\Delta r$. $\bR_i$ denotes the range channel and is part of $\bX_i$. $\mathbbm{1}$ is the indicator function and has the value 1 if the expression is true and 0 otherwise. $\delta$ is an indicator function based on the validity of each the participating pixels, defined as:
\begin{align}
\delta = \bM_i[m', n'] \cdot \bM_i[m, n]
\end{align}
$\delta$ also appears in subsequent kernels.

While $f_{2D+}$ takes the range information into account, it is very inefficient in that the number of parameters increases by $K$-fold, which can be significant and cause overfitting. Moreover, the amount of computation also increases by $K$-fold.
 
\PAR{Self-attention kernel.} Given the sparse nature of the range image data in the 3D space, graph operation are a more natural choice than projecting to a higher-dimensional space. The transformer \cite{vaswani2017attention} is one of the most popular graph operators. It has found success in both NLP \cite{devlin2018bert} and computer vision \cite{carion2020end}. In its core, the transformer generates weights depending on the input features and spatial locations of the features, and therefore does not require a set of weights in a dense form. A transformer-inspired kernel looks like follows:
\begin{align}
f_{SA} \coloneqq \sum_{m', n'} \text{softmax}( & \bF_i[m, n]^T \cdot \bW_q^T \cdot (\bW_k \cdot \bF_i[m', n'] + r))  \nonumber \\
& \cdot \bW_v \cdot \bF_i[m', n']  \cdot \delta 
\end{align}
\begin{align}
r = \bW_r \gamma(\bX_i[m, n], \bX_i[m', n']) \nonumber
\end{align}
where $\bW_q$, $\bW_k$, $\bW_v$ and $\bW_r$ are four sets of trainable weights. $\gamma(., .)$ is an \textbf{asymmetric positional encoding} between two points. It is defined as:
\begin{align}
\gamma(\bmx, \bmx') \coloneqq \{ &  r' \cdot \text{cos}(\Delta{\theta}) \cdot\text{cos}(\Delta{\phi}) - r,  \nonumber \\
& r' \cdot\text{cos}(\Delta{\theta}) \cdot  \text{sin}(\Delta{\phi}) , \nonumber \\
& r' \cdot \text{sin}(\Delta{\theta}) \}
\end{align}
\begin{align}
\Delta{\theta} = \theta' - \theta,  \quad   \Delta{\phi} = \phi' - \phi \nonumber
\label{eq:g}
\end{align}
where $\bmx=\{\theta, \phi, r\}$, $\bmx'=\{\theta', \phi', r'\}$ are the azimuth, inclination and depth of the points. $\gamma(.,.)$ is also used in subsequent kernels.

This asymmetric positional encoding has a geometric meaning. Namely, it is in an oblique Cartesian frame viewed from the sensor's location. For each pixel, after rotating the sphere by $-\theta$ and $-\phi$, $\bmx$ has the spherical polar coordinates $\{0, 0, r\}$, while $\bmx’$ is at $\{\Delta{\theta}, \Delta{\phi}, r’\}$. We project them to Cartesian, which yields $\{r, 0, 0\}$ for $\bmx$ and $\{\text{cos}(\Delta{\theta}) \cdot \text{cos}(\Delta{\phi}) \cdot r’, \text{cos}(\Delta{\theta})\cdot \text{sin}(\Delta{\phi}) \cdot r’, \text{sin}(\Delta{\theta}) \cdot r’\}$ for $\bmx’$. The encoding is then their element-wise difference. Note that the oblique Cartesian frame is different from pixel to pixel, but does not depend on the weights of each layer, and therefore can be pre-computed once for all layers per sample. This positional encoding is also used for the subsequent kernels.


\PAR{PointNet kernel.} While the transformer has seen great success in NLP and computer vision, PointNet \cite{qi2017pointnet} on the other hand has laid the groundwork to the majority of works for 3D point cloud understanding in the past years. It is widely used in robotics, thanks to VoxelNet \cite{zhou2018voxelnet}, PointPillars \cite{lang2019pointpillars} and PointRCNN \cite{shi2019pointrcnn}. The PointNet formulation is yet quite simple. It learns a multi-layer perceptron (MLP) that encodes the neighboring features and their relative coordinates to the center, and pools the encodings via max-pooling. Our PointNet-inspired kernel looks as follows:
\begin{align}
f_{PN} \coloneqq & \max_{m', n'} \text{MLP}( \nonumber \\
& [\bF_i[m', n'], \gamma(\bX_i[m, n], \bX_i[m', n'])], \Theta) \cdot \delta 
\end{align}
where $\Theta$ are trainable weights for the MLP. 

\PAR{EdgeConv kernel.} The edge convolution proposed by \cite{wang2019dynamic} is very similar to PointNet. In PointNet, the input to the MLP is the feature itself and a relative positional encoding. The edge convolution adds one more feature that is the center feature to the input:
\begin{align}
f_{EC} &\coloneqq \max_{m', n'} \text{MLP}( \nonumber \\
& [\bF_i[m', n'], \bF_i[m, n], \gamma(\bX_i[m, n], \bX_i[m', n'])], \Theta) \cdot \delta 
\end{align}

Although the last three, the Transformer, the PointNet, and EdgeConv kernels, are inspired by the 3D graph literature discussed in \secref{sec:Related Work}, they do not result in the inability to model occlusion and the inefficiency due to the need of the nearest neighbor search for each point. The perspective point-set aggregation layer can model occlusion just like any 2D range image-based method. Moreover, it does not require the nearest neighbor search, as it selects neighbors based on the distances in the 2D range image rather than in 3D. Finding neighbors in a dense 2D grid is trivial. 

\subsection{Smart Down-Sampling}
\label{sec:Smart Down- And Upsampling}

Unlike RGB images, the LiDAR range or RGBD images can have a noticeable amount of invalid range pixels in the image. It can be due to light-absorbing or less reflective surfaces. In the case of LiDAR, quantization and calibration artifacts can even result in missing returns that form a regular pattern, where the down-sampling with a fixed stride can inadvertently further emphasize the missing returns. Therefore, we define a smart down-sampling strategy to avoid missing returns as we sample: When we down-sample with a stride of, for example, 2$\times$2, we select 1 pixel from 4 neighboring pixels. But instead of always selecting the first or the last pixel, we select a \textit{valid} pixel, if available, which is the closest to the centroid of all valid pixels among the four. 


We define the down-sampling layer as follows (for brevity, we depict the math for the 1D space here):
\begin{align}
\bF_o[m] = \bF_i[\hat{m}],\; \bX_o[m] = \bX_i[\hat{m}],\; \bM_o[m] = \bM_i[\hat{m}] \\
\hat{m} = \argmin_{m' \in \setS} \| \bR_i[m'] - \mu \|^2,\; \mu =  \frac{\sum_{m' \in \setS} \bR_i[m']}{\| \setS \|_0} \nonumber
\end{align}
where $\setS$ contains valid $s' \in m \cdot \lambda, \ldots, (m+1) \cdot \lambda - 1$ according to the mask $\bM_i$, and $\lambda$ is the intended stride. $\bR_i$ is the range part of the spherical polar coordinates $\bX_i$. 

During up-sampling, technically, we would need to generate new points $\bX_o$ from the input 3D coordinates $\bX_i$, which is difficult to do. Luckily, an up-sampling usually mirrors a previous down-sampling and never exceeds the original input resolution. Therefore, we can remember the coordinates and the mask from the input of a corresponding down-sampling layer and reuse them for after up-sampling. We use the zeros vector as features for the new pixels generated from up-sampling. The up-sampling layer is the reverse operation of the down-sampling layer:
\begin{align}
\bX_o =& \bX_{i'},\; \bM_o = \bM_{i'} \nonumber \\
\bF_o[m] =&
\begin{cases}
\bF_i[\bar{m}],\; \text{if $m=\hat{m}$ and $m \in \setS$} \\
0,\; \text{else}
\end{cases} \\
\hat{m} =& \argmin_{m' \in \setS} \| \bR_o[m'] - \mu \|^2,\; \mu =  \frac{\sum_{m' \in \setS} \bR_o[m']}{\| \setS \|_0} \nonumber
\end{align}
where $\setS$ contains all valid $s' \in \bar{m} \cdot \lambda, \ldots, (\bar{m}+1) \cdot \lambda - 1$ according to the mask $\bM_{i'}$ where $\lambda$ is the up-sampling stride. $\bX_{i'}$ and $\bM_{i'}$ are values taken from a previous layer $i'$ whose down-sampling with stride $S$ yields the input layer $i$. $\bR_o$ is the range part of the spherical polar coordinates $\bX_o$.

\subsection{Backbone Architecture}
\label{sec:Backbone}

Now that we have both the perspective point-set aggregation layers and the sampling layers defined, we look at the backbone architecture to chain the layers together into a network. We performed a low-effort manual architecture search using the 2D convolution kernel and kept using the best architecture for the remaining experiments. Our network builds on top of the building blocks proposed in \cite{meyer2019lasernet}: the \textbf{feature extractor (FE)} that extracts features with an optional down-sampling, and the \textbf{feature aggregator (FA)} that merges most features to lower-level features to create skip connections. Our pedestrian network consists of 4 FE and 1 FA blocks, and predictions are made on half of the input resolution. Since the vehicles appear wider in the range image, we extend the network to 8 FE and 5 FA blocks. Please find an illustration of the architecture in \supsecref{sec:Perspective Range Image Backbone}.


\subsection{Point-Cloud-Camera Sensor Fusion}
\label{sec:Point-Cloud-Camera Sensor Fusion}

The perspective range image representation provides a natural way to fuse camera features to point cloud features since each location in the range image can be projected into the camera space. For each camera image, we first compute dense features using a modern convolution U-network \cite{ronneberger2015u} (please see \supsecref{sec:Camera Backbone and Fusion Details} for details). We project it to a location in its corresponding camera image for each point in the range image. We then collect the feature vector computed at that pixel in the camera image and concatenate the feature vector to the range image features. A zeros feature vector is appended, should an area be not covered by any camera. This approach can apply to any layer since there is always a point associated with each location. We train our networks end-to-end, with the camera convolution networks randomly initialized.

\subsection{CenterNet Detector}
\label{sec:CenterNet Detector}

We validate our point cloud representation via 3D object detection. We extend the CenterNet \cite{zhou2019objects} to 3D: For each pixel in the backbone network's output feature map, we predict both a classification distribution and a regression vector. The classification distribution contains $C+1$ targets, where $C$ is the number of classes plus a background class.

During training, the classification target is controlled by a Gaussian ball around the center of each box (see \supfigref{fig:centernet}): $s_{i,j} = \mathcal{N}(||\bmx_i - \bmb_j||_2, \sigma)$, where $\bmx$ are points and $\bmb$ are boxes. $\sigma$ is the Gaussian standard deviation, set to 0.25 meters for pedestrians and 0.5 meters for vehicles. For 2D detection in images, where CenterNet was originally proposed, the box center is always a valid pixel in the 2D image. In 3D, however, points are sparse, and the closest point to the center might be far away. Therefore, we normalize the target score by the highest score within each box to ensure that there is at least one point with a score 1.0 per box. We then take the maximum over all boxes and get the final training target score per point: $\bmy^{cls}_i = \max_j s_{i,j} / \max_{i \in B_j} s_{i,j}$.

The regression target $\bmy_i^{reg}$ for 3D detection contains 8 targets for 7 degrees-of-freedom boxes: a three-dimensional relative displacement vector from the point's 3D location to the center of the predicted box; another three dimensions that contain the absolute length, width, and height; and a single angle split into its sine and cosine forms in order to avoid discontinuity around $2\pi$.

We used the penalty-reduced focal loss for the classification, as proposed by \cite{zhou2019objects}, and the $\lone$-loss for the regression. We then train with a batch size of 256 over 300 epochs with an Adam optimizer. The initial learning rate is set to 0.001, and it decays exponentially over the 300 epochs.

All of our experiments applies the CenterNet detector head to the final feature map of the backbone. We have observed no significant difference in quality between CenterNet and other single-shot detectors such as the SSD \cite{liu2016ssd}. Two-stage methods, such as \cite{shi2019pointrcnn}, usually outperform single-stage methods on vehicles by a significant margin. However, the impact on pedestrians is less prevalent.




\section{Experiments}

\begin{table*}[!tbp]
\captionsetup{font=small}
\centering
\begin{tabular}{lcccccccc}
\toprule
\multirow{2}{*}{Method} & \multicolumn{2}{c}{3D} & \multicolumn{2}{c}{BEV} & \multicolumn{3}{c}{3D APH$_{L2}$ by distance} \\
& AP$_{L1}$ & \textbf{APH$_{L2}$} & AP$_{L1}$ & APH$_{L2}$ & <30m & 30-50m & >50m \\
\midrule       
LaserNet CVPR'19 \cite{meyer2019lasernet}* & 62.9 & 45.4 & 69.7 & 50.4 & 62.6 & 39.2 & 17.4 \\
PointPillars CVPR'19 \cite{lang2019pointpillars}* & 61.6 & 43.0 & 70.4 & 49.5 & 54.2 & 40.7 & 25.7 \\
MultiView CORL'19 \cite{zhou2019end} & 65.3 & - &  74.4 & - & - & - &  - \\
StarNet Arxiv'19 \cite{ngiam2019starnet} & 68.3 & 52.8 & 73.8 & 57.3 & 63.1 & 52.1 & 35.5 \\
PPBA-StarNet ECCV'20 \cite{cheng2020improving} & 69.7 & 53.9 & 74.9 & 58.4 & 64.4 & 53.2 & 36.6 \\
Pilar-based ECCV'20 \cite{wang2020pillar} & 72.5 & - & 78.5 & - & - & - & - \\
\midrule
\thismodel+ Conv2D (Ours) & 63.4 & 47.0 & 71.3 & 53.1 & 63.4 & 42.0 & 19.0 \\
\thismodel + RQ-Conv2D (Ours) & 68.4 & 54.3 & 76.9 & 61.6 & 67.1 & 52.6 & 32.2 \\
\thismodel + Self-Attention (Ours) & 57.9 & 43.7 & 65.3 & 49.5 & 60.3 & 39.8 & 17.1 \\
\thismodel + PointNet (Ours) & 72.4 & 57.9 & 79.3 & 63.9 & 70.3 & 56.3 & 35.5 \\
\thismodel+ EdgeConv (Ours) & 73.9 & 59.6 & 80.6 & 65.6 & 71.5 & 58.4 & 38.1 \\
\thismodel+ EdgeConv + Camera (Ours) & \textbf{75.5} & \textbf{61.5} & \textbf{82.2} & \textbf{67.6} & \textbf{72.3} & \textbf{61.3} & \textbf{41.3} \\
\bottomrule
\end{tabular}
\caption{\textbf{Pedestrians} on the Waymo Open Dataset \textit{validation} set. \textbf{3D APH$_{L2}$} is the primary metric for the dataset. Results denoted with * are based on our reimplementation. Others are taken from papers or via email communication with paper authors. Our method significantly improves over recent works. }
\label{tbl:peds}
\end{table*}

\begin{table*}[!tbp]
\captionsetup{font=small}
\centering
\begin{tabular}{lcccccccc}
\toprule
\multirow{2}{*}{Method} & \multicolumn{2}{c}{3D} & \multicolumn{2}{c}{BEV} & \multicolumn{3}{c}{3D APH$_{L2}$ by distance} \\
& AP$_{L1}$ & \textbf{APH$_{L2}$} & AP$_{L1}$ & APH$_{L2}$ & <30m & 30-50m & >50m \\
\midrule       
LaserNet CVPR'19 \cite{meyer2019lasernet}* & 56.1 & 48.4 & 73.1 & 63.9 & 75.1 & 45.6 & 21.7 \\
PointPillars CVPR'19 \cite{lang2019pointpillars}* & 56.1 & 48.2 & 77.2 & 67.6 & - & - & - \\
MultiView CORL'19 \cite{zhou2019end} & 62.9 & - & 80.4 & - & - & - & -  \\
StarNet Arxiv'19 \cite{ngiam2019starnet} & 55.1 & 48.3 & 67.7 & 60.0 & 79.1 & 43.1 & 20.2 \\
PV-RCNN CVPR'20 \cite{shi2020pv}$^\dagger$  & \textbf{70.3} & \textbf{64.8} & \textbf{83.0} & \textbf{67.6} & \textbf{91.0} & \textbf{64.5} & \textbf{35.7} \\
PPBA-PointPillars ECCV'20 \cite{cheng2020improving} & 61.8 & 53.4 & 81.4 & 72.2 & - & - & - \\
LSTM ECCV'20 \cite{huang2020lstm} & 63.4 & - & - & - & - & - & -  \\
Pillar-based ECCV'20 \cite{wang2020pillar}$^\dagger$  & 67.7 & -  & 86.1 & - & - & - & -\\
RCN CORL'20 \cite{Bewley2020RangeCD}  & 69.5 & -  & 83.4 & - & - & - & - \\
\midrule
\thismodel+ Conv2D (Ours) & 60.3 & 52.2 & 78.1 & 68.9 & 79.7 & 49.0 & 24.3 \\
\thismodel + RQ-Conv2D (Ours) & 56.8 & 49.2 & 76.2 & 67.2 & 75.7 & 46.1 & 22.8 \\
\thismodel + PointNet (Ours) & 64.5 & 56.2 & 80.5 & 71.6 & 80.7 & 54.6 & 31.2 \\
\thismodel+ EdgeConv (Ours) & 65.2 & 56.7 & 80.8 & 71.8 & 81.4 & 55.1 & 31.2 \\
\bottomrule
\end{tabular}
\caption{\textbf{Vehicles} on the Waymo Open Dataset \textit{validation} set. \textbf{3D APH$_{L2}$} is the primary metric for the dataset. Results denoted with * are based on our reimplementation. Others are taken from papers or via email communication with paper authors. Top two results per column are marked bold. Our method performs better than most published methods except for PV-RCNN \cite{shi2020pv}. $^\dagger$ PV-RCNN and RCN relies on a two-stage detection pipeline, and is therefore superior quality but less efficient than the other models in this table.}
\label{tbl:cars}
\end{table*}

\subsection{Waymo Open Dataset}

We conducted experiments on the pedestrians and vehicles of the Waymo Open Dataset \cite{sun2020scalability}. The dataset contains 1000 sequences, split into 780 training, 120 validation, and 100 test. Each sequence contains 200 frames, where each frame captures the full 360 degrees around the ego-vehicle that results in a range image of a dimension 64$\times$2650 pixels. The LiDAR has a maximum range of around 100 meters.

\PAR{Metrics.} We use metrics defined by the Waymo Open Dataset. \textbf{AP}: Average precision at 0.5 IOU for vehicles and 0.7 IOU for pedestrians. \textbf{APH}: Same as AP, but also takes the heading into account when matching boxes. \textbf{3D} vs. \textbf{BEV}: Whether IOU is measured on rotated 3D boxes or projected top-down rotated 2D boxes. \textbf{L1} vs. \textbf{L2}: Level of difficulty, L2 include more difficult boxes. 

Results for pedestrians and vehicles detections are highlighted in \tblref{tbl:peds} and \tblref{tbl:cars}. \figref{fig:samples} shows a few example results. Our method significantly outperforms all recent works on the pedestrian category, including the 3D grid or graph representations methods. This boost is not surprising for two reasons: a) pedestrians are tall so that the perspective view captures its full shape, b) they are also thin so that the voxels in the voxel-based methods end up too large and therefore cannot accurately make predictions.
We perform very competitively on vehicles, outperforming recently published methods, including several published this year.


\subsection{Detailed Kernel Analysis}
\label{sec:Detailed Analysis for Point-Set Aggregation Kernels}

\begin{figure}[tb]
\captionsetup{font=small}
  \centering
  \begin{subfigure}[b]{0.47\textwidth}
    \includegraphics[width=\textwidth]{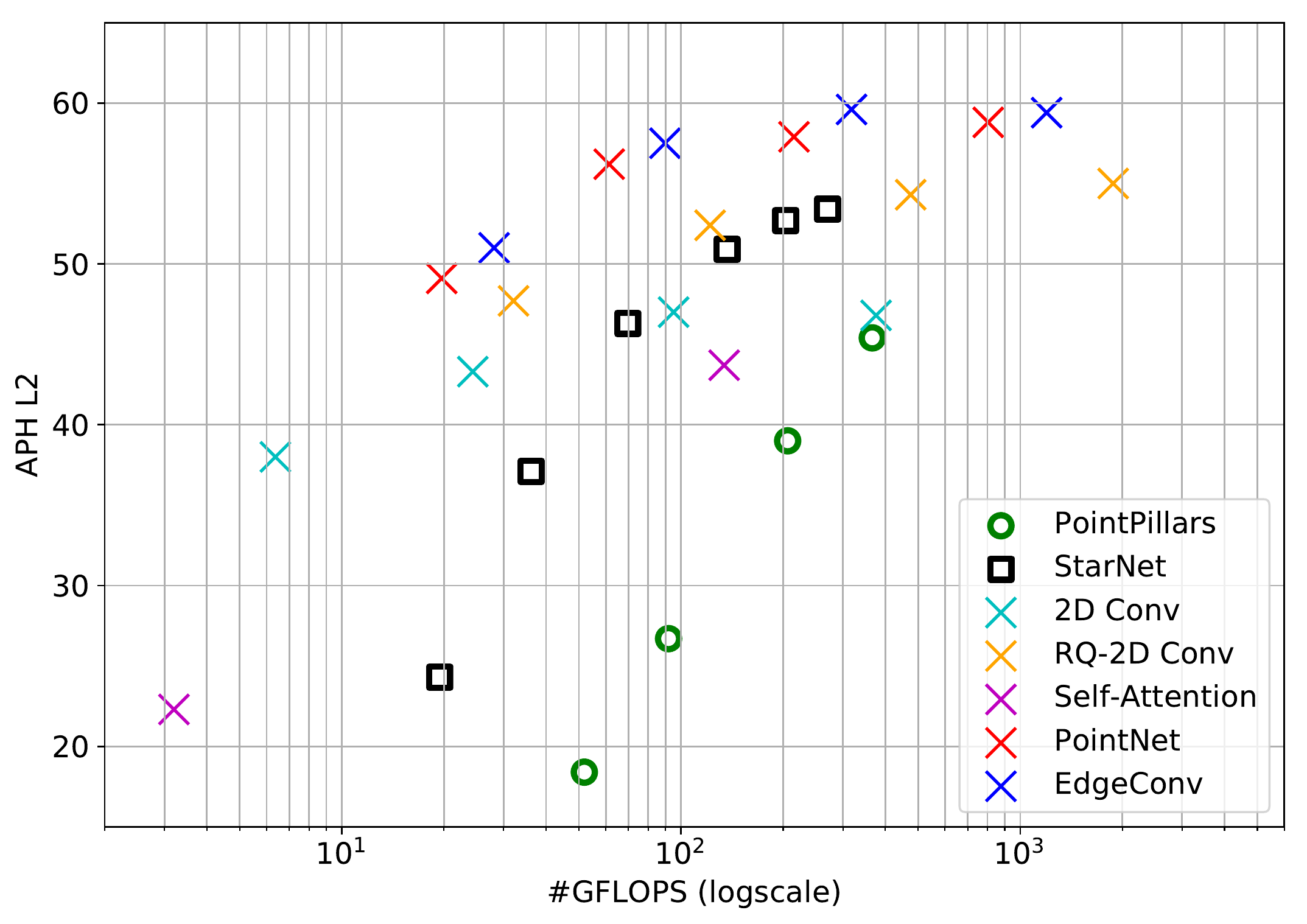}
    \caption{Complexity vs. accuracy}
    \label{fig:flops}
  \end{subfigure}
  \begin{subfigure}[b]{0.47\textwidth}
    \includegraphics[width=\textwidth]{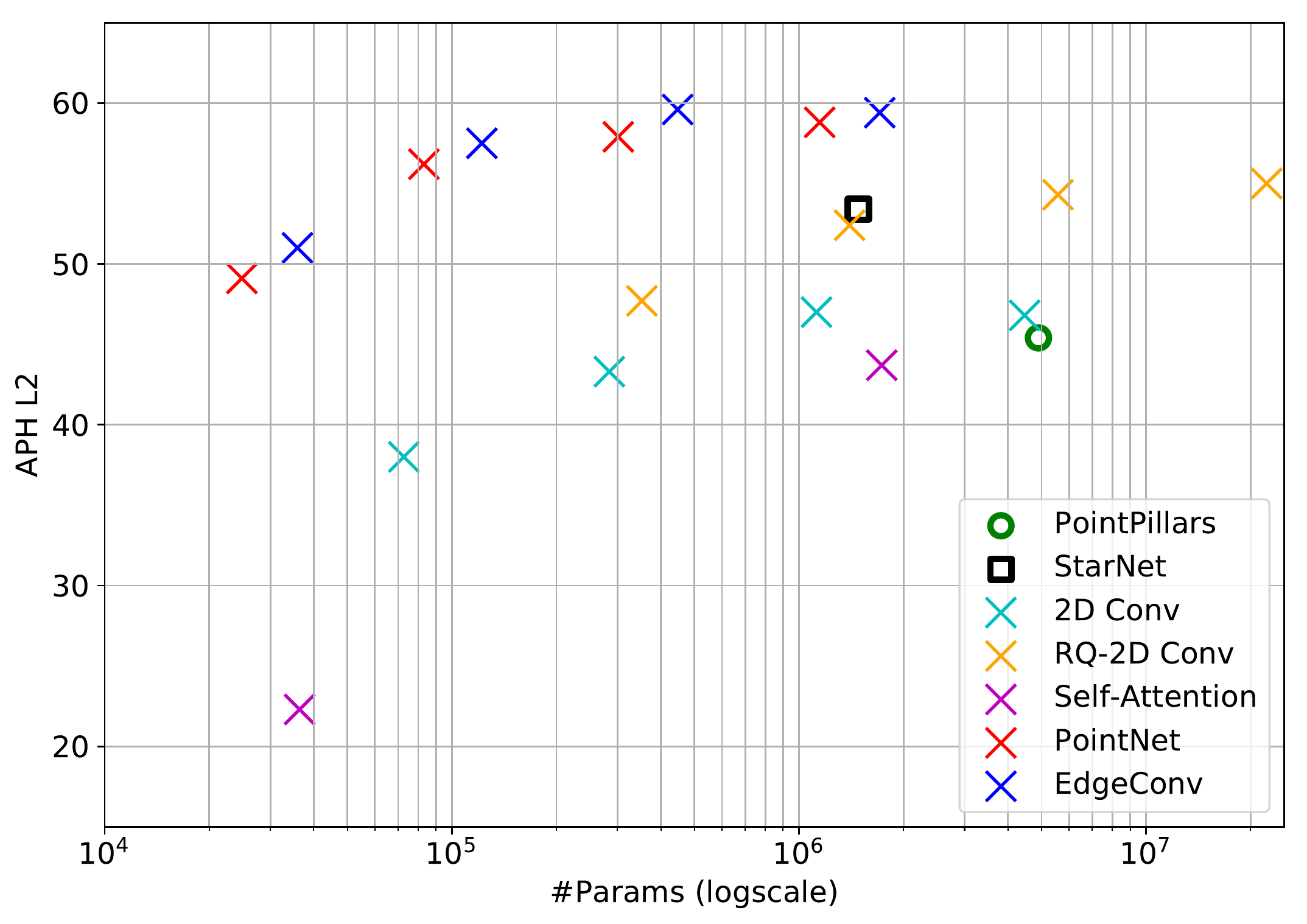}
    \caption{\#Parameters vs. accuracy}
    \label{fig:params}
  \end{subfigure}
\caption{Detailed analysis of the different Point-Set Aggregation Kernels. For each kernel, we train up to 4 models with a depth multiplier ranging from 0.25 to 2.0. PointPillars \cite{lang2019pointpillars} and StarNet \cite{ngiam2019starnet} are also plotted for reference. The operating points for PointPillars and StarNet are from one single model each, but with different input dimensions. a) Complexity vs. accuracy. b) Model size vs. accuracy. Experiments done on the Waymo Open Dataset pedestrians \textit{validation} set. The kernels with the best accuracy are PointNet and EdgeConv, while the most efficient kernels are the baseline 2D convolution and the transformer. We shed some light to the sub-par quality of the transformer in \secref{sec:Detailed Analysis for Point-Set Aggregation Kernels}. Note that the most efficient PointNet kernel outperforms PointPillars in quality, while needing \textit{180 times} fewer FLOPS and model parameters.
}
\end{figure}

\begin{table}[!tbp]
\captionsetup{font=small}
\centering
\begin{tabular}{ccccccc}
\toprule
EdgeConv & FE$_1$ & FE$_2$ & FE$_3$ & FA & FE$_4$ & Conv2D \\ 
\midrule          
59.6 & 56.8 & 54.9 & 51.7 &  53.2 & 55.2 & 47.0 \\
\bottomrule
\end{tabular}
\caption{Ablation to determine where the EdgeConv kernels are most effective. We start with a network of only 2D Conv kernels, and replace any one of the backbone blocks with EdgeConv. Experiments were done on the pedestrians of Waymo Open Dataset \textit{validation} set. The APH$_{L2}$ metric is reported.
}
\label{tbl:placement}
    \end{table}

\begin{table}[!tbp]
\captionsetup{font=small}
\centering
\begin{tabular}{cc}
\toprule
EdgeConv baseline & 59.6 \\
\midrule          
W/o smart down-sampling & 58.2 \\
Cartesian instead of polar & 54.2 \\
\bottomrule
\end{tabular}
    \caption{Ablation on the polar vs. Cartesian parametrization and the smart down-sampling strategy. Experiments done on the pedestrians of the Waymo Open Dataset \textit{validation} set. The APH$_{L2}$ metric is reported.
    }
    \label{tbl:ablation}
\end{table}

We take a closer look to compare the five kernels introduced in \secref{sec:Point-Set Aggregation Kernels}. Since different kernels have different computational complexity, it is unfair to compare them by their quality alone. \figref{fig:flops} shows a complexity-vs.-accuracy analysis, while \figref{fig:params} shows the model-size-vs.-accuracy analysis. For each kernel, we train multiple models with each with a different depth multiplier, ranging from 0.25 to 2.0. A depth multiplier is a factor that is applied to the number of channels in each layer in a network and is a simple way to yield multiple models at different complexity and accuracy.

\PAR{Complexity vs. accuracy.} The \textbf{2D} kernel baseline is one of the least expensive methods, as expected. The \textbf{Range-Quantized 2D} kernel with quantization of 4 buckets adds approximately four times to the complexity and improves significantly over the baseline. \textbf{Self-attention} is fairly cheap in computation. However, it does not perform as well as any other kernel. We believe that the reason might be that we kept the kernel size to be 3$\times$3 throughout the network, and the transformer works better with larger kernel sizes. \textbf{PointNet} and \textbf{EdgeConv} have a relatively small difference in quality at around 1-2\% at 20-30\% difference in computation. 


\PAR{Model size vs. accuracy.} \thismodel has a significant advantage over existing models in terms of model size. Our smallest model, a PointNet kernel model with a depth multiplier of 0.25, achieves higher accuracy than the baseline PointPillars, while only having \textbf{24k} parameters compared to close to 5M for PointPillars.


\subsection{Mix-and-Match Kernels}

\begin{figure*}[tb]
\captionsetup{font=small}
  \centering
\includegraphics[width=\textwidth]{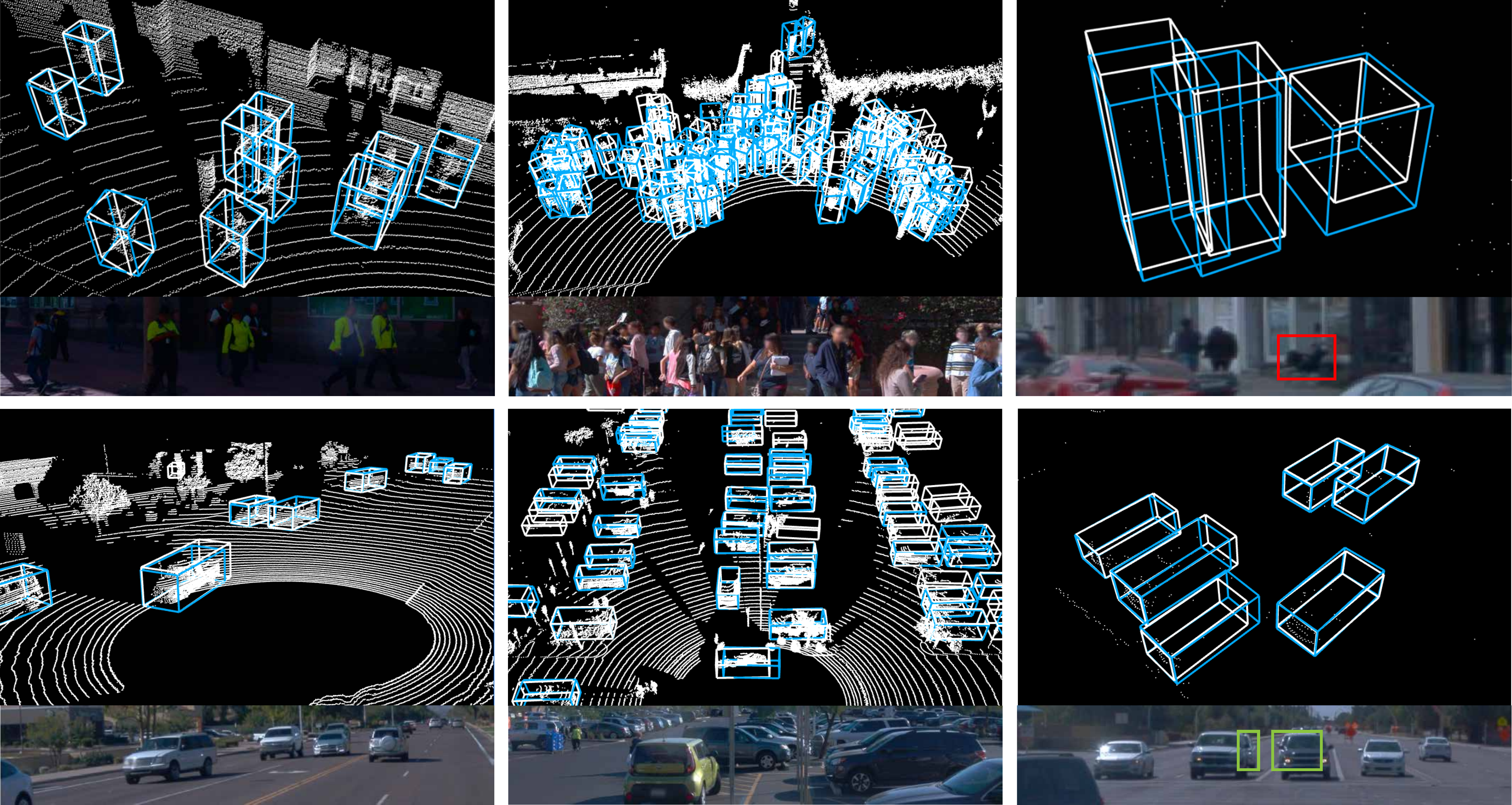}
\caption{(Best viewed in color) Example pedestrian and vehicle detection results of \thismodel + EdgeConv on the Waymo Open Dataset. White boxes are groundtruth and \textcolor{blue}{blue} boxes are our results. \textbf{Left}: Our method performs well when objects are close and mostly visible. \textbf{Center}: It can also handle large crowds with severe occlusion. Many of the false-negatives in the center bottom image have no points in the groundtruth boxes. \textbf{Right}: It can also detect objects in the long range where points become sparse. Note in the top right image, the pedestrian on the right (highlighted in a \textcolor{red}{red} box on the image) is sitting in a chair. And in the bottom right example, the there is severe occlusion (\textcolor{green}{green} boxes) for the two cars behind the front two.
\vspace{-0.4cm}
}
\label{fig:samples}

\end{figure*}

In the previous section, we noticed that a network consisting of all EdgeConv kernels delivers the strongest results. However, we also observed that the EdgeConv kernel with the same depth multiplier is not as efficient as the 2D kernel. In this ablation, we study if we can keep most of the 2D kernels while only applying the EdgeConv in a few layers. \tblref{tbl:placement} shows the accuracy numbers. \textbf{Conv2D} and \textbf{EdgeConv} are networks of only 2D or EdgeConv kernels and serve as a pair of pseudo upper and lower bounds. We then replace each of the blocks of the backbone from 2D to EdgeConv. Interestingly, replacing either the first or the last block generates the most benefits. Since our backbone resembles a U-Net \cite{ronneberger2015u}, it means that the EdgeConv kernel has the most impact with larger resolutions.

\subsection{Additional Ablation}

\PAR{Point-Cloud-Camera Sensor Fusion.} Each frame in the Waymo Open Dataset comes with five calibrated camera images capturing views from the front and sides of the car. We downsize each camera image to $400\times600$ pixels and use a convolutional neural network to extract a 192-dimensional feature at each location.  These features are concatenated to one of the layers in the perspective point cloud network.  We experimented with fusing the features at different layers of the network and found that the model performed better regardless of the layer we chose to fuse. Our best result in \tblref{tbl:peds} is obtained when the camera features fuses at the input to the first extractor layer.

\PAR{Sensor Polar vs. World Cartesian.} In \eqref{eq:g}, the asymmetric positional encoding is defined in the spherical polar coordinate system around the sensor. It is possible to take the displacement vector between two points in the projected world Cartesian frame instead. If so, the method's overall concept becomes very similar to PointNet++ \cite{qi2017pointnet++}, with the main difference being that the neighbors of a point are taken from the perspective range image grid rather than through nearest neighbor search. Operating in the polar coordinate system is natural in the range image. However, the Cartesian system has a strong prior in the heading since most objects move perpendicular to the ego-vehicle. In \tblref{tbl:ablation} we show the ablation of polar vs. Cartesian for the \thismodel + EdgeConv model on pedestrians. It appears that the benefits of operating in the polar coordinate system significantly outweigh the drawbacks at 59.6\% vs. 54.2\% APH$_{L2}$.

\PAR{Smart Down-Sampling.} The smart down- and up-sampling strategy outlined in \secref{sec:Smart Down- And Upsampling} allows the down-sampling to avoid missing returns at the cost that the down-sampling no longer follows a regular pattern. As shown in \tblref{tbl:ablation}, the smart sampling technique yields 1.4\% benefit in APH$_{L2}$.

\section{Conclusion and Limitations}
\label{sec:Conclusion}

This paper presents a new 3D representation based on the range image, which leverages recent advances in graph convolutions. It is efficient and yet powerful, as demonstrated on pedestrians and vehicles on the Waymo Open Dataset.

It is not without limitations. Most 3D detection tasks use a 7 degrees-of-freedom that only has a yaw rotation around the Z-axis in the world coordinate system. Suppose the sensor has a significant pitch or roll wrt. the world coordinate system, the boxes no longer appear only yaw-rotated in the range image. It is an issue for indoor scene datasets but less of a problem for autonomous driving configurations, where the rotating LiDAR usually sits upright to the world coordinate system.

Another challenge is data augmentation. \cite{cheng2020improving} in \tblref{tbl:peds} and \tblref{tbl:cars} shows a significant improvement by applying data augmentation to PointPillars \cite{lang2019pointpillars} and StarNet \cite{ngiam2019starnet}. In 3D, data augmentation can be diverse and effective. When points are in the dense range image form, we can no longer apply most of them without disturbing the dense structure. We also observed that the EdgeConv kernel network is not sensitive to strategies that are still reasonable in the range image, e.g., random flip and random points drop.


{\small
\bibliographystyle{ieee_fullname}
\bibliography{myrefs}
}

\clearpage

\newpage

\appendix






\section{Additional Details on the Backbone}
\label{sec:Perspective Range Image Backbone}

We use the basic building blocks proposed in \cite{meyer2019lasernet}: the \textbf{feature extractor (FE)} and the \textbf{feature aggregator (FA)}. Figure 4 in \cite{meyer2019lasernet} shows a detailed diagram. In words, the FE block consists of ten 3$\times$3 filters. Every two layers are grouped and bypassed by a skip connection. Whenever a straightforward skip connection is not possible due to mismatched spatial resolution or depth, a 1$\times$1 filter with potential striding is applied. The FA block is used to up-sample lower resolution feature maps back to a high resolution for skip connections. It first applies a transposed convolution filter to up-sample the lower resolution feature, which is concatenated to a skip connection from a high-resolution feature from a previous layer before down-sampling. The combined feature then undergoes 4 additional 3$\times$3 filters.

\begin{figure*}[h]
\captionsetup{font=small}
  \begin{subfigure}[b]{\textwidth}
    \centering
    \includegraphics[width=0.6\textwidth]{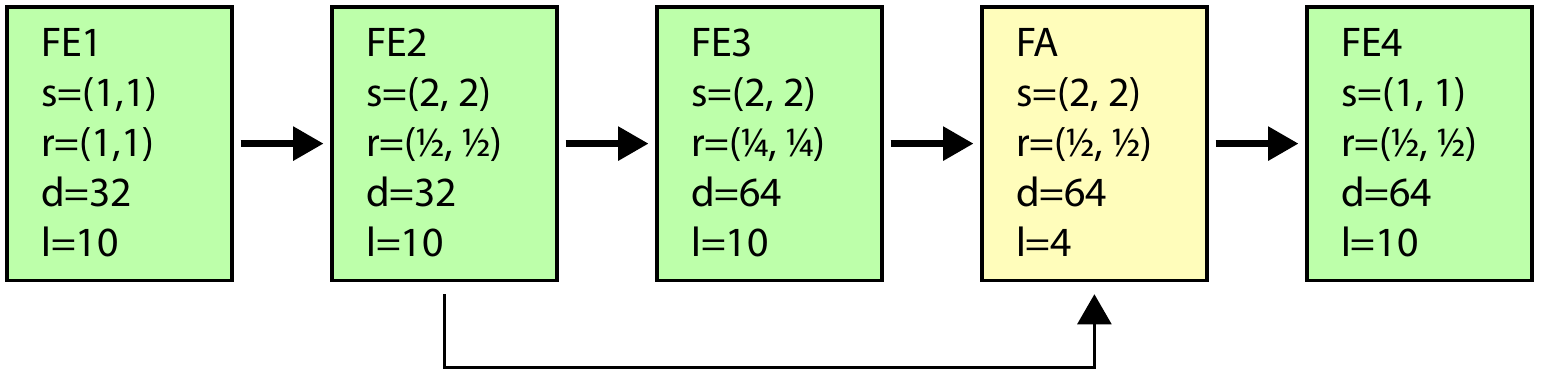}
    \caption{Backbone for pedestrians. \vspace{0.4cm}}
  \end{subfigure}
  \begin{subfigure}[b]{\textwidth}
    \centering
    \includegraphics[width=0.8\textwidth]{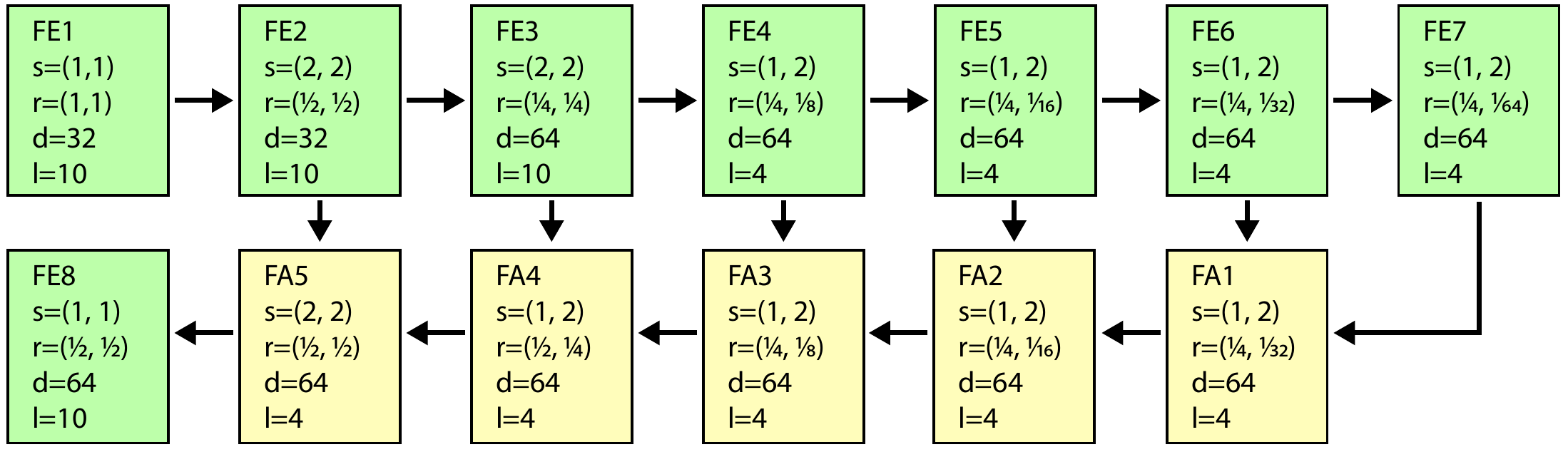}
    \caption{Backbone for vehicles.}
  \end{subfigure}
\caption{Backbone architecture for 3D pedestrian and vehicle detection. FE and FA are feature extractors and feature aggregators outlined in the text. \textbf{s} is stride in height and width and is applied at the start of each block. \textbf{r} is the relative resolution to the original input size after applying the stride. \textbf{d} is the number of channels for all convolutional layers inside the block. \textbf{l} is the number of 3$\times$3 convolutional layers.}
\label{fig:backbone}
\end{figure*}

\figref{fig:backbone} shows backbone architectures for pedestrians and vehicles. Vehicles appear wider in the range image and require a larger receptive field. Therefore, the vehicle model is a lot deeper. The new feature extractors only have 4 instead of 10 convolutional layers each.

\section{Additional Details on the Camera Backbone and Fusion}
\label{sec:Camera Backbone and Fusion Details}

We use a U-Net \cite{ronneberger2015u} that resembles the 2D backbone network as depicted in Figure 4 of \cite{zhou2018voxelnet}. We make a small modification in that we skip the initial down-sampling by a stride of 2. The network has 16 convolutional layers with a kernel of 3$\times$3 each, split into 3 blocks at an increasingly smaller resolution. Features from each of the 3 blocks are concatenated to create the final feature map.



\begin{figure*}[h]
\captionsetup{font=small}
  \centering
\includegraphics[width=\textwidth]{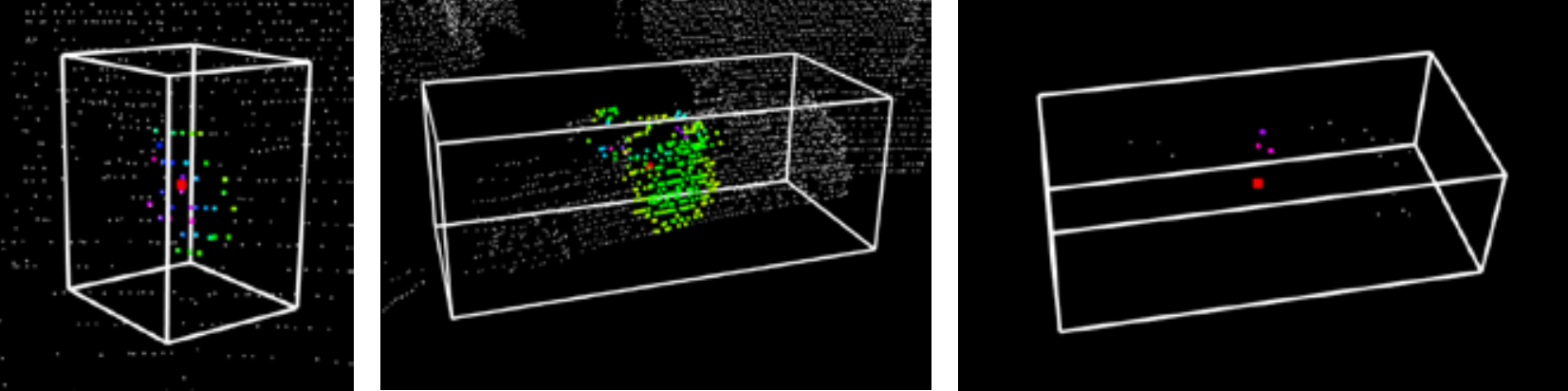}
\caption{3D CenterNet classification targets. White box denotes the groundtruth bounding box. Red dot is the center of the box. White points have the target value 0.0. Rainbow colors indicate the target values ranging from 0.1 (green) to 1.0 (purple). Left: pedestrian. Center: a close-by car with dense points. Right: a far away car with sparse points. Note that the closest points have the target value 1.0 despite being relatively far away from the center due to normalization.}
\label{fig:centernet}
\end{figure*}

\end{document}